\documentclass[11pt]{article}
\usepackage{fontspec}  
\PassOptionsToPackage{hyphens}{url}  
\usepackage[preprint]{acl}
\usepackage{times}
\usepackage{latexsym}
\usepackage[T1]{fontenc}
\usepackage[utf8]{inputenc}
\usepackage{microtype}
\usepackage{inconsolata}
\usepackage{booktabs}
\usepackage{array}
\newcolumntype{L}[1]{>{\raggedright\arraybackslash}p{#1}}  
\usepackage{graphicx}
\usepackage{amsmath}
\usepackage{placeins}
\usepackage{flafter}
\usepackage{float}
\usepackage{tikz}
\usetikzlibrary{arrows.meta,fit,positioning}
\newfontfamily\devanagarifont{NotoSansDevanagari-Regular.ttf}[Path=fonts/, Script=Devanagari, Scale=MatchLowercase]
\newfontfamily\telugufont{NotoSansTelugu-Regular.ttf}[Path=fonts/, Script=Telugu, Scale=MatchLowercase]
\newenvironment{planitems}
  {\begin{itemize}\setlength{\itemsep}{1pt}\setlength{\parskip}{0pt}\setlength{\parsep}{0pt}}
  {\end{itemize}}
\newenvironment{exampleitems}
  {\raggedright\begin{itemize}\setlength{\itemsep}{1pt}\setlength{\parskip}{0pt}\setlength{\parsep}{0pt}}
  {\end{itemize}}
\newenvironment{sourceitems}
  {\small\raggedright\begin{itemize}\setlength{\itemsep}{1pt}\setlength{\parskip}{0pt}\setlength{\parsep}{0pt}}
  {\end{itemize}}

\title{Broad and Efficient PII Detection with Unified Labels}
\author{Andy Hsu and Konstantin Berlin \\ Cisco AI Defense \\ \texttt{\{andhsu, berlink\}@cisco.com}}

\begin{document}
\raggedbottom
\maketitle

\begin{abstract}
Personally Identifiable Information (PII) detection models must efficiently find identifiers across diverse contexts while maintaining a low false-positive rate (FPR) on high-volume traffic. Recent detectors increasingly rely on complex architectures that raise deployment cost, yet many are fine-tuned on only a few synthetic datasets that limit their coverage. Combining public datasets is challenging because their taxonomies differ, and even after categories are mapped to a common schema, labels can disagree on what counts as a detection, producing inconsistent training and evaluation signals. To address these problems, we present a practical 141M-parameter encoder-based detector trained on 1.12 million examples from 77 public datasets, relabeled under a detailed definition of broad contextual detection for 12 structured PII categories. Evaluated on the AI4Privacy and Nemotron-PII benchmarks at a low-FPR operating point, our model achieves the highest recall against larger baselines in 11 of 14 comparisons under our labels and 9 of 14 under the original labels. Its largest recall gains include useful additional coverage, such as multilingual disclosures and IP addresses embedded in URLs. It achieves this with 2--10$\times$ higher throughput and 94--97\% less peak GPU memory than the baselines, delivering leading detection performance at substantially lower cost.

\end{abstract}

\section{Introduction}\label{sec:1}

\begin{figure}[t]
  \centering
  \includegraphics[width=\columnwidth]{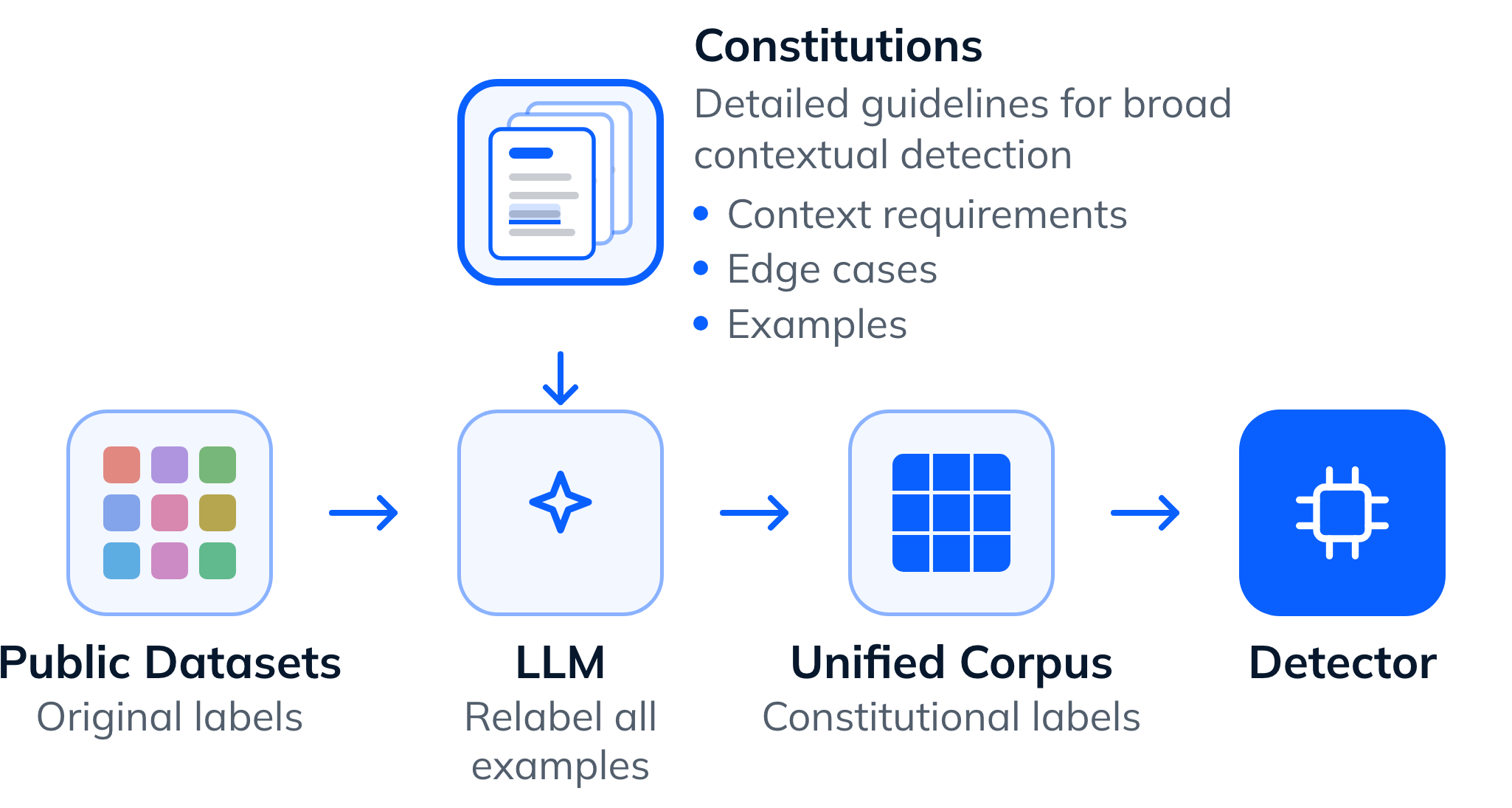}
  \caption{\textbf{Approach overview.} A single set of contextual PII detection criteria is used to relabel various datasets, combine them into a single large dataset, and train a compact model on the unified labels.}
  \label{fig:method-overview}
\end{figure}

Personally Identifiable Information (PII) detectors are a key part of data loss prevention (DLP) systems and privacy guardrails, which often screen large volumes of traffic under tight latency budgets \cite{alneyadi2016dlp,asthana2025oneshield}. They must surface identifiers with high recall at a low false-positive rate (FPR) so that downstream components can decide which are sensitive and how to handle them \cite{schwartz2011pii,nist800122}. In the past, detectors largely relied on hand-written deterministic rules such as regular expressions combined with keyword checks, which can both overflag and generalize poorly \cite{meystre2010deidentification,presidio,dernoncourt2017deidentification,subramani2023detecting}. Machine learning models were introduced to provide more robust contextual detection and are now widely used \cite{dernoncourt2017deidentification}. Even so, recent evaluations find that current PII detection models still fall short of the coverage and precision needed in practice \cite{zhang2026piitrace,uppala2026opf}. A critical factor in performance is the data and labels used for training: a practical model needs positive examples that cover a wide range of formats and languages, as well as diverse negative examples to avoid overflagging real-world traffic.

Despite this, many PII models today are still fine-tuned on one or a few datasets that are synthetic and identifier-heavy \cite{piiranha,nvidia_gliner_pii}. Adding more datasets to training can be challenging because their taxonomies differ widely and are hard to combine at scale. Even after schemas are aligned, the merged labels within a category can disagree on what counts as a detection \cite{yang2025b2ner,openai_privacy_filter}, producing an inconsistent training and evaluation signal. For example, they may disagree on whether a credit card number should be labeled when only the last four digits are exposed, or whether a business phone number should be labeled at all. At the same time, newer releases are moving to more complex architectures to support flexible category coverage \cite{nvidia_gliner_pii,zaratiana2026gliner2pii} or more nuanced judgments such as whether an entity is private \cite{openai_privacy_filter}. These models are harder to deploy and raise cost and latency, all without solving the underlying problem of what they are trained on.

Our approach focuses on addressing the data and labeling problems in order to build a better model. First, we clearly define how we want our PII detector to behave. Beyond deciding which categories to cover, we write a detailed definition for each category that defines the key aspects of how its contextual detection should operate. These definitions are broad across countries, formats, and languages, so the detector surfaces a wide range of useful identifiers. Second, we completely relabel datasets using our definitions, replacing the decisions made by different annotators and aligning the labels at the example level (Figure~\ref{fig:method-overview}). Large language models (LLMs) have opened up a scalable way to do this, and prior work shows that detailed category specifications can lead to more accurate and consistent labeling \cite{sainz2024gollie,berlin2026constitutional}. This relabeling process can be easily applied to any dataset and also allows us to incorporate examples from benign distributions that did not originally have PII labels. Training on the resulting data, we then find that a small model is not only far more efficient than larger baselines, but can also outperform them in recall at low FPR. This allows it to serve as a first-stage scanner for all traffic, so more expensive detection stages only need to run on a smaller set of candidates. Specifically, our contributions are:

\begin{planitems}
  \item We write contextual definitions of PII for 12 core identifier categories that guardrails commonly monitor. We then use GPT-5 Mini to directly apply our definitions and relabel two PII benchmark datasets. Compared with the original labels, the resulting labels remove many positive examples that lack context words for the category and add meaningful coverage such as identifiers embedded within larger spans.
  \item Using the same pipeline, we relabel 77 public datasets into a unified training corpus of approximately 1.12M examples.
  \item We train a compact, highly efficient 141M-parameter detector that produces both category detections and spans in a single inference pass, and evaluate it against larger models. At a low-FPR operating point, it achieves the highest recall in 11 of 14 comparisons under our labels and 9 of 14 under the benchmarks' original labels, which it is never trained on, while delivering 2--10$\times$ higher throughput and using 94--97\% less peak GPU memory.
\end{planitems}

\section{Related Work}\label{sec:2}

\subsection{PII Datasets}\label{sec:2.1}

PII datasets are highly diverse and typically include the positions of identifiers in the text as span labels in order to support span extraction tasks. Some examples include domain-specific resources such as the i2b2/UTHealth corpus and TAB \cite{stubbs2015i2b2,pilan2022tab}, multilingual corpora such as MAPA \cite{arranz2022mapa}, and synthetic resources such as AI4Privacy, Nemotron-PII, and SPY \cite{ai4privacy_open_pii_masking_500k,nemotron_pii,savkin2025spy}. This variety offers a wide range of training and evaluation examples, but the datasets are hard to combine or compare because their taxonomies differ in category names, breadth, and granularity.

Prior work on combining datasets with span labels mainly focuses on aligning taxonomies. In general named entity recognition (NER), B\textsuperscript{2}NER combines 54 datasets into a single training set by mapping their labels to a universal taxonomy of over 400 entity types, splitting and renaming types whose definitions conflict across datasets \cite{yang2025b2ner}. For PII specifically, PIIBench maps more than 80 labels from ten public datasets into 48 canonical categories \cite{jha2026piibench}. However, aligning taxonomies does not change the labels themselves, so differences remain in how each dataset labeled its examples. We go a step further and relabel every example using a single definition.

\subsection{Labels and Labeling Definitions}\label{sec:2.2}

LLMs are increasingly being used to produce training signals for NER and PII models. UniversalNER distills an LLM into smaller NER models \cite{zhou2024universalner}, and NuNER pre-trains a compact encoder on LLM-labeled text \cite{bogdanov2024nuner}. Detailed written guidelines have been shown to help LLMs follow labeling conventions more accurately and consistently \cite{sainz2024gollie,berlin2026constitutional}. For PII, concurrent work uses frontier LLMs to label multilingual assistant conversations \cite{zhang2026piitrace}. These efforts primarily focus on previously unlabeled text from a single source. Our approach instead labels many different datasets for a diverse distribution, replacing the original labels of those that already covered our PII categories.

Prior work has found that benchmarks that appear to be for the same NLP task can follow different underlying methodologies in how they label examples \cite{subramonian-etal-2023-takes}, and this also holds for PII detection. For example, the SPY benchmark labels identifiers as PII only when they refer to the author of the text \cite{savkin2025spy}. Benchmark labels can also contain errors that distort evaluation \cite{northcutt2021pervasive,rucker2023cleanconll}. OpenAI Privacy Filter's model card argues that missing labels in AI4Privacy PII-Masking-300k understate its performance: an LLM reviewer validated \textasciitilde{}10,000 spans the model detected but the labels omitted, and excluding them from the errors raised its reported F1 score \cite{openai_privacy_filter}. Our analysis relabels two complete benchmarks, identifies recurring disagreement patterns independently of detector predictions, then observes their effects across multiple models.

\subsection{PII Detection Models}\label{sec:2.3}

Traditional pattern-based PII detectors use regular expressions, dictionaries, and context rules to recognize known identifier forms \cite{meystre2010deidentification,presidio,subramani2023detecting}. Machine-learning models instead use the surrounding text as context and typically return each identifier as a span \cite{dernoncourt2017deidentification,zaratiana2023gliner}. Fixed-category detectors are often fine-tuned on one or a few synthetic datasets; Piiranha, for example, is trained on AI4Privacy PII-Masking-400k \cite{piiranha}.

PII detectors have also started to adopt more flexible task formulations. The GLiNER family lets users define the categories at inference time by conditioning span predictions on label names supplied with each input \cite{zaratiana2023gliner}, and GLiNER2 extends this to labels with natural-language descriptions, along with classification and structured extraction \cite{zaratiana2025gliner2}. Both have been adapted specifically for PII detection \cite{nvidia_gliner_pii,zaratiana2026gliner2pii}, and popular options use uni-encoders that encode the label definitions together with every input. As a result, specifying the intended behavior adds inference cost to all traffic and this cost grows as the behavior is specified in more detail. GLiNER Guard avoids this cost with a bi-encoder variant that encodes labels separately and caches them, but its deployed pipeline still detects structured identifiers such as phone numbers and tax identifiers with rule-based components \cite{minko2026glinerguard}. Another line of work encodes a model's own definition of what is private into its weights and judges whether each disclosed entity is private by that definition. Privacy Filter and the concurrent PII-Tracer both adapt decoder language models into bidirectional classifiers that tag privacy-oriented categories with constrained BIOES decoding \cite{openai_privacy_filter,zhang2026piitrace}.

The capabilities of these newer architectures can be useful, but all of them still depend on the data and task definition they are trained on. When what a model has learned does not match an application's requirements, it can require further training. For Privacy Filter, analysis has shown that a fine-tuned XLM-RoBERTa encoder \cite{conneau2020xlmr} surpasses its zero-shot performance with only 100 to 1,000 labeled examples, depending on the domain \cite{uppala2026opf}.

\section{Method}\label{sec:3}

\subsection{Task Definition}\label{sec:3.1}

Our taxonomy covers 12 structured identifier categories that appear widely across PII taxonomies: Bank Account Number, Bank Routing Code, Credit Card Number, Driver License Number, Email Address, IP Address, Medical License Number, Passport Number, Patient ID Number, Phone Number, Social Security Number, and Tax Identification Number. Given an input text, the detector identifies every category that is present and also extracts each identifier as a span. An input can contain multiple categories or identifiers. We define the task broadly: surface every identifier whose category is clear from the context. Applications can then decide which identifiers matter for their use case.

A span contains the identifier value but excludes the surrounding words that describe it. For example, in \texttt{My credit card number is 4111 1111 1111 1111.}, the extracted span is \texttt{4111 1111 1111 1111}. Spans assigned to different categories may also overlap when one value serves several roles in context, such as a U.S. Social Security number used as a tax identifier.

How much evidence is needed depends on the category. An Email Address has a distinctive structure, so its format alone is often enough. On the other hand, an ambiguous digit sequence needs supporting context before it is labeled as a Bank Account Number, such as mentions of banking or payments. Rule-based DLP systems implement a similar principle by confirming a pattern match with nearby context words \cite{presidio,microsoft_purview_sit}.

We also do not judge whether a contextually supported identifier has a valid format before flagging it. For example, a value presented as a Credit Card Number remains positive even if it fails a checksum. Validity rules such as checksums and issuer formats vary by country and can change over time, so deterministic checks can apply them more reliably downstream. This also preserves detections that an application may still want to review, such as partially masked numbers.

\subsection{Constitutions and Data}\label{sec:3.2}

We write a constitution for each of our 12 categories that defines what belongs in the category, what evidence is required, which text belongs in the span, and how common edge cases should be handled. Appendix~\ref{app:A} provides an example of the guidance included in each constitution. For each input, GPT-5 Mini \cite{openai2025gpt5developers} receives the text and the constitutions for several categories in a single labeling prompt. It returns labels for both of our task outputs in a structured format, together with a rationale for each decision: input-level labels for the categories present in the text and span labels for each identifier. During development, we inspect samples of the output and refine the constitutions when we find recurring deviations. We call the resulting annotations constitutional labels, and Section~\ref{sec:4.1} analyzes how they differ from the original labels of our evaluation datasets.

The training corpus contains approximately 1.12M relabeled examples drawn from 77 public datasets released under permissive licenses (Appendix~\ref{app:G}). To provide varied benign contexts, it includes a wide range of AI and security-related text such as chatbot conversations, jailbreak prompts, and agent traces. It also includes many PII datasets with diverse identifier formats and contexts. For evaluation, we relabel the validation split of AI4Privacy Open PII Masking 500k \cite{ai4privacy_open_pii_masking_500k} and the test split of NVIDIA Nemotron-PII \cite{nemotron_pii}. We refer to these as the AI4Privacy and Nemotron datasets moving forward. None of the examples used for evaluation are used to train our model.

\subsection{Model}\label{sec:3.3}

\begin{figure}[t]
  \centering
  \includegraphics[width=\columnwidth]{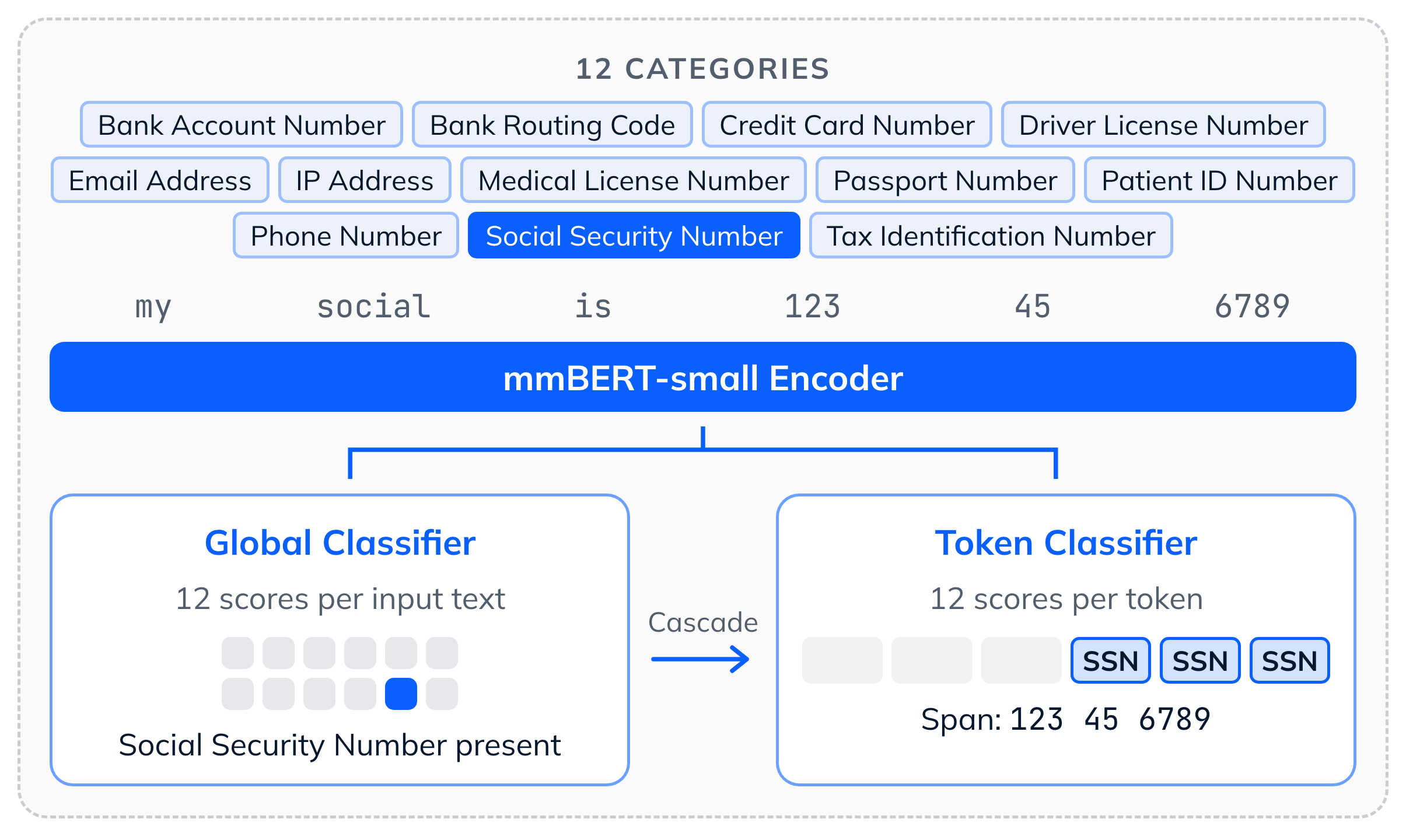}
  \caption{\textbf{Model architecture.} Two classifiers share one mmBERT-small encoder in a single forward pass. The global classifier detects categories present in the input and serves as a prefilter for the token classifier, which extracts spans for categories that the global classifier flagged.}
  \label{fig:model-architecture}
\end{figure}

We use mmBERT-small \cite{marone2026mmbert} as our base model because it combines recent encoder training methods with large-scale multilingual pretraining in a compact, conventional architecture suited to fast inference. We fine-tune this approximately 141M-parameter model on the unified, constitutionally labeled corpus, adding two classifiers to the shared encoder (Figure~\ref{fig:model-architecture}). The global classifier is trained on input-level labels and predicts which categories are present in the whole input. The token classifier is trained on span labels and scores all 12 categories for every token, with consecutive tokens above the threshold forming spans.

Similar to joint intent detection and slot filling \cite{liu2016joint}, we train both classifiers together, which improved span extraction recall compared to training the token classifier alone. We believe this is because the global classifier learns a simpler task: recognizing that an identifier and its supporting context are present in the input is easier than extracting the identifier's exact span. At inference, the two classifiers act as a cascade: we keep spans from the token classifier only for categories that the global classifier flags. The global classifier filters out values with similar formats that the token classifier could otherwise overflag at a lower threshold, so the cascade can use that lower threshold to recover more spans. Training takes only 25 minutes on four NVIDIA RTX PRO 6000 GPUs and Appendix~\ref{app:B} provides further training details. We also found during development that increasing model size did not translate to significant performance gains: training the same recipe on the larger mmBERT-base (307M parameters) improved average recall by less than one percentage point while increasing memory and latency.

\section{Results}\label{sec:4}

\subsection{Label Disagreement Analysis}\label{sec:4.1}

\begin{figure*}[t]
  \centering
  \includegraphics[width=0.96\textwidth]{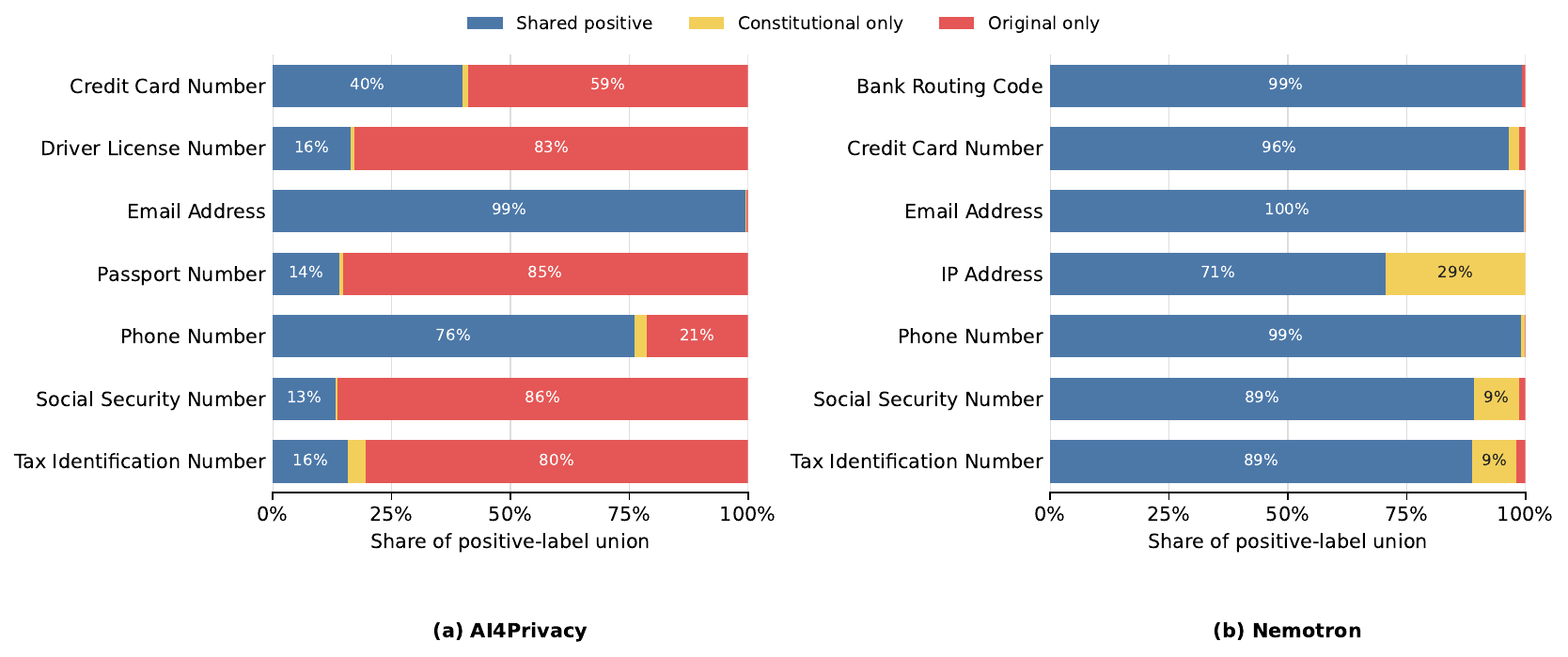}
  \caption{\textbf{Paired label composition.} Distribution of positive examples across the constitutional and original labels for each comparable category. Each row partitions the union of positives into examples labeled by both sets, only the constitutional labels, or only the dataset's original labels. In five AI4Privacy categories, 59--86\% of positives are original-only, while Nemotron mostly agrees except for constitutional-only IP Address examples.}
  \label{fig:label-composition}
\end{figure*}

We first compare our constitutional labels against the original labels of the AI4Privacy and Nemotron datasets to see where they differ, and find that the two benchmarks show very different disagreement patterns (Figure~\ref{fig:label-composition}). For each category, we define shared positives as the examples labeled positive by both label sets.

\begin{table}[t]
  \centering\small
  \setlength{\tabcolsep}{3pt}
  \begin{tabular*}{\columnwidth}{@{\extracolsep{\fill}}lcc@{}}
    \toprule
    Category & Direct context & Broad context \\
    \midrule
    Credit Card Number & 29.3\% / 0.3\% & 71.2\% / 32.5\% \\
    Driver License Number & 83.3\% / 0.3\% & 96.1\% / 31.6\% \\
    Passport Number & 88.1\% / 0.2\% & 92.0\% / 27.3\% \\
    Social Security Number & 62.3\% / 0.5\% & 87.4\% / 24.5\% \\
    Tax Identification Number & 36.3\% / 0.0\% & 68.7\% / 24.1\% \\
    \midrule
    Combined & 54.2\% / 0.24\% & 80.4\% / 27.5\% \\
    \bottomrule
  \end{tabular*}
  \caption{\textbf{Context word presence in AI4Privacy original labels.} Percentage of examples containing context words anywhere in the text. Each cell reports shared / original-only positives. Broad context also counts the category's direct words.}
  \label{tab:1}
\end{table}

\paragraph{AI4Privacy Disagreement.}
On AI4Privacy, our labels mostly remove positives from the original labels. The largest disagreements occur in five categories (Credit Card Number, Driver License Number, Passport Number, Social Security Number, and Tax Identification Number), where 59--86\% of positives appear only in the original labels. Looking more closely at these removed positives, we find that many do not give sufficient contextual evidence for their category. For example, \texttt{Don't forget to bring your YV9022888!} is labeled as containing a Passport Number, but nothing in the sentence indicates a passport.

To measure how common this is, we search each text for two types of context words. Direct words name the category, such as \texttt{passport} or \texttt{SSN}, and broad words give looser related context, such as \texttt{ID} or \texttt{payment}. Appendix~\ref{app:C} shows the full list of context words, which include translations for all eight languages in AI4Privacy. The results show that direct context words appear anywhere in the text for 54.2\% of shared positives but only 0.24\% of original-only positives. Broad context words show a similarly large gap, at 80.4\% and 27.5\% (Table~\ref{tab:1}). These findings indicate that our labels mainly remove positives that lack context words for their category, as our definition intends.

\paragraph{Nemotron Disagreement.}
In contrast, our labels mostly agree with Nemotron's original labels and mainly add positives where they differ. The largest difference is in IP Address, where 29\% of positives appear only in our labels. Nearly all of these added positives (99.6\%) are IP addresses inside URLs. For example, Nemotron labels \nolinkurl{http://192.168.1.100:8080/biometric/results/match_notice.html} as a URL but does not separately label \texttt{192.168.1.100} as an IP Address. The constitutional labels mark the embedded value as an IP Address because its context sufficiently identifies the category, showing that our labels can also add positives that the original labels leave out.

\paragraph{Label Audit.}
As an independent evaluation, we sample 50 disagreements per category from AI4Privacy positives in the five major categories that our labels remove and Nemotron IP addresses that our labels add. Two LLM judges from different model families, GPT-5.6-Sol and Claude Opus 5, decide whether the context supports that each category is present \cite{zheng2023judging}. On AI4Privacy, GPT-5.6-Sol agrees with our labels on 68\% of the sampled disagreements and Claude Opus 5 on 72\%, averaged over the five categories. On Nemotron, they agree with our labels on 100\% and 98\% of the sampled IP addresses. A manual review of the cases where both judges agree with the original labels finds that most are either inconclusive, such as ``Could you please send me a copy of your S2469568 so I can complete your registration?'' (translated from Italian), or rely on less common country-specific formats as the only evidence, such as ``Can you give me your 756.0754.6599.08 so I can help you with your request?'' (translated from French), which matches the format of a Swiss social security number. We believe recognition of these international formats as sufficient context can be improved by adjusting the constitutions, which we leave to future work.

\subsection{Detection Performance}\label{sec:4.2}

\paragraph{Category Mapping.}
We compare our detector with three recent models that report high benchmark performance and are licensed for commercial use, ranging from 307M to 1.4B total parameters: NVIDIA GLiNER \cite{nvidia_gliner_pii}, Fastino GLiNER2 \cite{zaratiana2026gliner2pii}, and OpenAI Privacy Filter \cite{openai_privacy_filter}. Because the benchmarks and models use different category names, we map their labels to our 12 categories, keeping only clear matches with similar scope. For example, \texttt{email} maps directly to Email Address, while \texttt{ipv4} and \texttt{ipv6} together cover IP Address. We reject broader or incomplete matches such as \texttt{account\_number}, which can also cover non-bank accounts. After mapping, seven categories on each benchmark can be compared with at least one baseline, giving 14 comparisons. For GLiNER models, we use only the label names documented as supported in each model's card or training data. Appendix~\ref{app:D} lists each mapping, and Appendix~\ref{app:E} describes how each model's category scores and spans are computed.

\paragraph{Operating Point.}
Detectors that perform well on PII benchmarks can still substantially overflag benign traffic \cite{zhang2026piitrace}. To evaluate the detectors under conditions closer to deployment, we set their thresholds to meet an input-level FPR constraint on 100,000 real user conversations from WildChat \cite{zhao2024wildchat}. We define input-level FPR as the percentage of conversations where a detector flags a category that does not appear anywhere in the text. Because WildChat does not have labels for our PII categories, we label the conversations with our constitutions to compute FPR. We also considered label-free calibration, which has issues that Appendix~\ref{app:F} discusses in more detail. For each detector and category, we select the lowest threshold with at most 0.1\% FPR. Appendix~\ref{app:F} reports results at looser limits of 0.25\% and 0.5\% FPR.

\paragraph{Findings.}
We first evaluate each model's ability to correctly detect presence of the category in the text, scoring the same outputs against both sets of labels (Table~\ref{tab:2}). Our model has the highest recall in 11 of 14 comparisons under the constitutional labels. Notably, our model is trained only on constitutional labels but also has the highest recall in 9 of 14 comparisons under the original labels.

\begin{table*}[t]
  \centering\small
  \begin{tabular*}{\textwidth}{@{\extracolsep{\fill}}llcccc@{}}
    \toprule
    Dataset & Category & Ours & NVIDIA GLiNER & Fastino GLiNER2 & Privacy Filter \\
    \midrule
    AI4Privacy & Credit Card Number & \textbf{0.976 / 0.955} & 0.685 / 0.573 & 0.568 / 0.416 & --- \\
     & Driver License Number & \textbf{0.909 / 0.306} & --- & 0.723 / 0.185 & --- \\
     & Email Address & \textbf{0.999 / 1.000} & 0.983 / 0.983 & 0.982 / 0.983 & 0.990 / 0.990 \\
     & Passport Number & \textbf{0.967 / 0.740} & --- & 0.825 / 0.197 & --- \\
     & Phone Number & 0.904 / 0.797 & 0.808 / 0.743 & 0.935 / 0.886 & \textbf{0.948 / 0.953} \\
     & Social Security Number & \textbf{0.923 / 0.237} & 0.494 / 0.126 & --- & --- \\
     & Tax Identification Number & \textbf{0.854 / 0.394} & 0.637 / 0.243 & 0.395 / 0.110 & --- \\
    Nemotron & Bank Routing Code & 0.999 / 0.994 & \textbf{1.000 / 0.996} & 0.628 / 0.623 & --- \\
     & Credit Card Number & \textbf{0.997} / 0.996 & \textbf{0.997 / 0.999} & 0.975 / 0.974 & --- \\
     & Email Address & 0.998 / 0.997 & \textbf{1.000 / 0.999} & 0.999 / \textbf{0.999} & 0.986 / 0.986 \\
     & IP Address & \textbf{0.986 / 0.999} & 0.721 / 0.995 & 0.652 / 0.922 & --- \\
     & Phone Number & \textbf{0.995 / 0.996} & 0.989 / 0.991 & 0.915 / 0.916 & 0.866 / 0.868 \\
     & Social Security Number & \textbf{0.991 / 0.994} & 0.979 / 0.989 & --- & --- \\
     & Tax Identification Number & \textbf{0.969} / 0.990 & 0.929 / \textbf{0.999} & 0.847 / 0.875 & --- \\
    \bottomrule
  \end{tabular*}
  \caption{\textbf{Detection recall at 0.1\% WildChat FPR.} Each cell reports constitutional / original recall on the same benchmark examples. Model outputs and thresholds remain unchanged between the two values. Recall is rounded to three decimal places, and the highest value in each comparison is bolded. A dash denotes that a comparison was not available under our methodology.}
  \label{tab:2}
\end{table*}

\begin{table*}[t]
\begin{minipage}[t]{0.48\textwidth}
  \centering\small
  \renewcommand{\thetable}{\arabic{table}a}\renewcommand{\theHtable}{\thetable}
  \begin{tabular*}{\linewidth}{@{\extracolsep{\fill}}lcc@{}}
    \toprule
    Language & Ours & NVIDIA GLiNER \\
    \midrule
    German & \textbf{0.989} & 0.615 \\
    English & \textbf{1.000} & 0.655 \\
    Spanish & \textbf{0.969} & 0.671 \\
    French & \textbf{0.983} & 0.631 \\
    Hindi & \textbf{0.653} & 0.411 \\
    Italian & \textbf{0.984} & 0.875 \\
    Dutch & \textbf{1.000} & 0.542 \\
    Telugu & \textbf{0.654} & 0.506 \\
    \bottomrule
  \end{tabular*}
  \caption{\textbf{Shared positive recall against NVIDIA GLiNER.} Recall on AI4Privacy positives shared by both label sets for Credit Card Number, Social Security Number, and Tax Identification Number. The higher value in each comparison is bolded.}
  \label{tab:3a}
\end{minipage}\hfill
\begin{minipage}[t]{0.48\textwidth}
  \centering\small
  \addtocounter{table}{-1}
  \renewcommand{\thetable}{\arabic{table}b}\renewcommand{\theHtable}{\thetable}
  \begin{tabular*}{\linewidth}{@{\extracolsep{\fill}}lcc@{}}
    \toprule
    Language & Ours & Fastino GLiNER2 \\
    \midrule
    German & \textbf{0.989} & 0.527 \\
    English & \textbf{0.994} & 0.681 \\
    Spanish & \textbf{0.976} & 0.667 \\
    French & \textbf{0.991} & 0.725 \\
    Hindi & \textbf{0.767} & 0.280 \\
    Italian & \textbf{0.987} & 0.781 \\
    Dutch & \textbf{1.000} & 0.622 \\
    Telugu & \textbf{0.674} & 0.258 \\
    \bottomrule
  \end{tabular*}
  \caption{\textbf{Shared positive recall against Fastino GLiNER2.} Recall on the same shared positives for Credit Card Number, Driver License Number, Passport Number, and Tax Identification Number.}
  \label{tab:3b}
\end{minipage}
\end{table*}

On AI4Privacy, our model leads the strongest comparable baseline by more than five percentage points under both label sets for Credit Card Number, Driver License Number, Passport Number, Social Security Number, and Tax Identification Number. To better understand what is contributing to the recall advantage, we examine the shared positives (examples labeled positive by both label sets) in these categories that our model detects but the baselines miss. We find that our model detects substantially more shared positives across all eight languages of the dataset (Tables~\ref{tab:3a} and~\ref{tab:3b}). For Nemotron, our model detects the embedded IP addresses observed in Section~\ref{sec:4.1}, which the GLiNER baselines largely miss.

We also observe that recall on AI4Privacy is generally lower across models under the original labels, dropping by up to 69 percentage points even though the evaluation data and model predictions did not change. Another recall difference can be seen in Nemotron IP Address, where our recall exceeds the GLiNER models' by 0.4 and 7.8 percentage points under the original labels but by 27 and 33 points under the constitutional labels. These shifts closely match the disagreement patterns seen in Section~\ref{sec:4.1}.

\subsection{Span Extraction}\label{sec:4.3}

We evaluate span extraction by computing exact-match micro F1 between each model's predicted spans and the constitutional span labels. Because each model splits text into tokens differently, we match spans by character position. A predicted span is correct only if it has the same category, start position, and end position as a labeled span.

Candidate thresholds come from the same WildChat calibration set used for detection, at per-category FPR targets from 0.025\% to 0.5\%. For both AI4Privacy and Nemotron, we split the evaluation data into a 20\% validation split and an 80\% test split. For each model and category, we choose the threshold with the best validation F1 and apply it to the test split. Our model achieves higher test F1 in five of the six benchmark--baseline pairs, including all three on Nemotron (Table~\ref{tab:4}).

\begin{table}[H]
  \centering\small
  \setlength{\tabcolsep}{2pt}
  \begin{tabular*}{\columnwidth}{@{\extracolsep{\fill}}llccc@{}}
    \toprule
    Dataset & Baseline & Categories & \shortstack{Ours\\F1} & \shortstack{Baseline\\F1} \\
    \midrule
    AI4Privacy & NVIDIA GLiNER & 5 & \textbf{0.865} & 0.791 \\
     & Fastino GLiNER2 & 6 & 0.816 & \textbf{0.851} \\
     & Privacy Filter & 2 & \textbf{0.895} & 0.840 \\
    Nemotron & NVIDIA GLiNER & 7 & \textbf{0.935} & 0.905 \\
     & Fastino GLiNER2 & 6 & \textbf{0.937} & 0.934 \\
     & Privacy Filter & 2 & \textbf{0.985} & 0.922 \\
    \bottomrule
  \end{tabular*}
  \caption{\textbf{Span extraction performance.} Exact-match micro F1 against the constitutional labels on each benchmark's test split. Categories counts the number of categories we compare between the two models, over which F1 is computed. The higher value in each comparison is bolded.}
  \label{tab:4}
\end{table}

\subsection{Inference Efficiency}\label{sec:4.4}

We measure inference cost on 600 WildChat inputs, split evenly among short, medium, and long examples, on one NVIDIA RTX PRO 6000 GPU in bfloat16 (Table~\ref{tab:5}). Measurements cover the full inference path, from tokenization to span decoding. Our model delivers 2--10$\times$ higher throughput and uses 94--97\% less peak GPU memory than the baselines.

\begin{table}[H]
  \centering\small
  \setlength{\tabcolsep}{2pt}
  \begin{tabular*}{\columnwidth}{@{\extracolsep{\fill}}lcccc@{}}
    \toprule
    Model & Params & Memory & Latency & Inputs/s \\
    \midrule
    Ours & 141M & \textbf{0.33} & \textbf{19.9} & \textbf{59.65} \\
    NVIDIA GLiNER & 445M & 11.03 & 76.5 & 13.43 \\
    Fastino GLiNER2 & 307M & 5.77 & 26.6 & 30.78 \\
    Privacy Filter & 1.4B / 50M & 6.25 & 173.9 & 6.05 \\
    \bottomrule
  \end{tabular*}
  \caption{\textbf{Inference efficiency.} Peak GPU memory is in GiB and median latency in milliseconds. Params counts all weights in the released checkpoint. For Privacy Filter, a mixture-of-experts model, the second value is the active parameters per token reported in its model card. Latency is measured with batch size 1 and throughput with batch size 8. The best value in each column is bolded.}
  \label{tab:5}
\end{table}

\section{Discussion}\label{sec:5}

Within the same FPR limit as the baselines, our model correctly flags many genuine disclosures that the baselines do not. On AI4Privacy, these are identifiers that both label sets agree on across all languages. When the baselines miss these identifiers, they usually label them as the wrong category, as in the examples below:

\begin{exampleitems}
  \item \textbf{Tax Identification Number (Hindi):} {\devanagarifont आपका आयकर नंबर} \texttt{88415.99447} {\devanagarifont है।} (\emph{\={a}pk\={a} \={a}ykar nambar 88415.99447 hai}; ``Your income tax number is 88415.99447.'') NVIDIA GLiNER labels it as Bank Routing Code.
  \item \textbf{Credit Card Number (German):} Bitte geben Sie uns Ihr \texttt{6518154446209804} an, um Ihre Diplom-Kosten zu bezahlen. (``Please give us your 6518154446209804 to pay your diploma fees.'') NVIDIA GLiNER labels it as Tax Identification Number and Fastino GLiNER2 as Phone Number.
  \item \textbf{Driver License Number (French):} Le numéro de permis de conduire de Sullton est \texttt{IP8QGH7J85}. (``Sullton's driver's license number is IP8QGH7J85.'') Fastino GLiNER2 labels it as IP Address.
\end{exampleitems}

This misrouting can have important consequences because guardrails often take actions with different severity based on the category assignment, such as monitoring phone numbers while blocking tax identification numbers. On Nemotron, the additional coverage is primarily embedded IP addresses. NVIDIA GLiNER, which was trained on Nemotron's training split, appears to inherit the dataset's omission of these cases.

Because our task is broad, not every contextual detection flagged by our model will matter to every application. For example, our model flags organizational phone numbers that OpenAI Privacy Filter does not under its \texttt{private\_phone} definition. Keeping these detections gives our model more flexibility when different applications want to adjust their coverage or evolve how they make sensitivity decisions, while a detector with a narrower or more subjective definition would likely require retraining \cite{openai_privacy_filter}. Delivering all of these qualities at low-FPR operating conditions and much lower cost gives our model strong practical value in real-world guardrail deployments.

\section{Conclusion}\label{sec:6}

In this paper, we presented a compact PII detector trained on many different public datasets relabeled under a single written definition of contextual detection, showing that it can outperform larger baselines at much lower cost. We plan to open source the model, inference code, example constitutions illustrating the detection criteria for each category, and sample relabeled data. We hope these artifacts support further research on which model sizes, architectures, and detection tasks best suit privacy guardrail applications. Future work will continue to refine the constitutions and extend coverage to additional identifier categories.

\section*{Limitations}

Our detector supports 12 structured identifier categories, but only 9 of them have comparable coverage in the selected benchmarks and baseline models. The remaining 3 require additional evaluation. As with any LLM, GPT-5 Mini may apply the constitutions incorrectly or inconsistently despite iterative review during constitution development. Operating points are calibrated on WildChat and evaluated on AI4Privacy and Nemotron. Other traffic distributions may require additional evaluation and recalibration.

\section*{Ethical Considerations}

All training and evaluation data come from public research datasets or synthetic sources. WildChat users affirmatively opted into anonymous research-data collection \cite{zhao2024wildchat}. We use these conversations only to calibrate aggregate false-positive rates and do not reproduce raw WildChat content. All qualitative examples shown in the paper come from synthetic benchmarks rather than WildChat.

Our detector is designed to help protect personal information by flagging identifiers so that they can be masked or reviewed. It is a classifier that only assigns category scores and spans to the text it is given and does not generate text, so it cannot reproduce content from its training data. Because it can still miss identifiers, it should be treated as one part of a larger privacy system rather than relied on alone.

Generative AI assistants supported paper editing and code development. The authors reviewed the resulting work and remain responsible for the accuracy, originality, and integrity of the paper and its accompanying artifacts.

\bibliography{main}

\clearpage

\appendix

\section{Constitution Examples}\label{app:A}

Table~\ref{tab:a1} shows how the Credit Card Number constitution covers each of the four aspects described in Section~\ref{sec:3.2}.

\begin{table}[H]
  \centering\small
  \renewcommand{\thetable}{A1}\renewcommand{\theHtable}{A1}
  \begin{tabular}{@{}L{0.21\columnwidth}L{0.73\columnwidth}@{}}
    \toprule
    Aspect & Guidance and example \\
    \midrule
    Scope & The primary account number or an explicitly identified partial segment of it, but not the CVV, CSC, or CVC. \texttt{My credit card number ends in 4327} $\rightarrow$ \texttt{4327} \\
    \addlinespace[3pt]
    Evidence & The text must identify the value as a card number, since matching the format is not enough. \texttt{Your order number is 4111 1111 1111 1111.} $\rightarrow$ not labeled \\
    \addlinespace[3pt]
    Span & The number itself, without the surrounding words that describe it. \texttt{The credit card number is 1111-2222-3333-4444} $\rightarrow$ \texttt{1111-2222-3333-4444} \\
    \addlinespace[3pt]
    Edge cases & A value presented as a card number is labeled even if it fails the checksum. \texttt{My credit card number is 1234-5678-9012-3456} $\rightarrow$ \texttt{1234-5678-9012-3456} \\
    \bottomrule
  \end{tabular}
  \caption{\textbf{Example constitutional guidance.} Guidance from the Credit Card Number constitution, with one example per aspect. Every constitution covers all four aspects.}
  \label{tab:a1}
\end{table}

\section{Training Configuration}\label{app:B}

We train the pretrained encoder and both classifiers together for one epoch and use the final checkpoint (Table~\ref{tab:b1}). Each classifier produces 12 independent category scores, so an input or token can be assigned to multiple categories.

\begin{table}[H]
  \centering\small
  \renewcommand{\thetable}{B1}\renewcommand{\theHtable}{B1}
  \begin{tabular}{@{}L{0.37\columnwidth}L{0.57\columnwidth}@{}}
    \toprule
    Setting & Value \\
    \midrule
    Base model & \nolinkurl{jhu-clsp/mmBERT-small} \\
    Training examples & 1,116,154 \\
    Chunk size / overlap & 512 / 128 tokens \\
    Epochs & 1 \\
    Batch size & 512 chunks (128 per GPU) \\
    Hardware & 4 $\times$ NVIDIA RTX PRO 6000 \\
    Training time & \textasciitilde{}25 minutes \\
    Precision & bfloat16 \\
    Optimizer & AdamW \\
    Learning rate & $2\times10^{-5}$, linear decay, no warm-up \\
    Weight decay & 0.01 \\
    Dropout & 0.1 before each classifier \\
    Loss & Unweighted sum of global and token binary cross-entropy \\
    \bottomrule
  \end{tabular}
  \caption{\textbf{Training configuration.} Configuration of the training run for our model.}
  \label{tab:b1}
\end{table}

\section{Context Words}\label{app:C}

We normalize each text before matching. Normalization applies Unicode NFKC, lowercases the text, and replaces every non-alphanumeric character with a space. A word or phrase matches only as a complete word sequence, so \texttt{id} does not match inside \texttt{valid}. Words from all eight languages are checked against every text. Direct words are specific to each category (Table~\ref{tab:c1}), while broad words are shared across categories (Table~\ref{tab:c2}).

\begin{table*}[tp]
  \centering\small
  \renewcommand{\thetable}{C1}\renewcommand{\theHtable}{C1}
  \begin{tabular}{@{}L{0.20\textwidth}L{0.08\textwidth}L{0.66\textwidth}@{}}
    \toprule
    Category & Language & Direct context words \\
    \midrule
    Credit Card Number & German & kreditkarte, kreditkartennummer, debitkarte, debitkartennummer, zahlungskarte, kartennummer \\
     & English & credit card, credit card number, debit card, debit card number, payment card, card number, card no \\
     & Spanish & tarjeta de crédito, tarjeta de credito, tarjeta de débito, tarjeta de debito, tarjeta bancaria, tarjeta de pago, número de tarjeta, numero de tarjeta \\
     & French & carte de crédit, carte de credit, carte de débit, carte de debit, carte bancaire, carte de paiement, numéro de carte, numero de carte \\
     & Hindi & {\devanagarifont क्रेडिट कार्ड}, {\devanagarifont डेबिट कार्ड}, {\devanagarifont भुगतान कार्ड}, {\devanagarifont कार्ड नंबर}, {\devanagarifont कार्ड संख्या} \\
     & Italian & carta di credito, carta di debito, carta bancaria, carta di pagamento, numero della carta, numero di carta \\
     & Dutch & creditcard, credit card, creditcardnummer, debetkaart, betaalkaart, kaartnummer \\
     & Telugu & {\telugufont క్రెడిట్ కార్డ్}, {\telugufont డెబిట్ కార్డ్}, {\telugufont చెల్లింపు కార్డ్}, {\telugufont కార్డ్ నంబర్}, {\telugufont కార్డ్ సంఖ్య} \\
    \addlinespace[3pt]
    Driver License Number & German & führerschein, fuhrerschein, führerscheinnummer, fuhrerscheinnummer, fahrerlaubnis, fahrerlaubnisnummer \\
     & English & driver license, driver s license, driving license, driving licence, driver licence, driver s licence, driving permit \\
     & Spanish & licencia de conducir, licencia de manejo, permiso de conducir, permiso de manejo, carnet de conducir, carné de conducir \\
     & French & permis de conduire, numéro de permis, numero de permis \\
     & Hindi & {\devanagarifont ड्राइविंग लाइसेंस}, {\devanagarifont ड्राइविंग लाइसेन्स}, {\devanagarifont चालक लाइसेंस} \\
     & Italian & patente di guida, patente \\
     & Dutch & rijbewijs, rijbewijsnummer \\
     & Telugu & {\telugufont డ్రైవింగ్ లైసెన్స్} \\
    \addlinespace[3pt]
    Passport Number & German & reisepass, reisepassnummer, passnummer \\
     & English & passport \\
     & Spanish & pasaporte \\
     & French & passeport \\
     & Hindi & {\devanagarifont पासपोर्ट} \\
     & Italian & passaporto \\
     & Dutch & paspoort, paspoortnummer \\
     & Telugu & {\telugufont పాస్ పోర్ట్}, {\telugufont పాస్}{\devanagarifont {\telugufont ‌}}{\telugufont పోర్ట్}, {\telugufont పాస్పోర్ట్} \\
    \addlinespace[3pt]
    Social Security Number & German & sozialversicherung, sozialversicherungsnummer, rentenversicherungsnummer \\
     & English & social security, social insurance, national insurance, ssn \\
     & Spanish & seguridad social, nss \\
     & French & sécurité sociale, securite sociale, nir \\
     & Hindi & {\devanagarifont सामाजिक सुरक्षा}, {\devanagarifont सोशल सिक्योरिटी}, {\devanagarifont एसएसएन} \\
     & Italian & previdenza sociale, sicurezza sociale \\
     & Dutch & burgerservicenummer, sociale zekerheid, bsn \\
     & Telugu & {\telugufont సామాజిక భద్రత}, {\telugufont సోషల్ సెక్యూరిటీ} \\
    \addlinespace[3pt]
    Tax Identification Number & German & steueridentifikationsnummer, steuer id, steuernummer, ust id, ust idnr, umsatzsteuer id, umsatzsteuer identifikationsnummer \\
     & English & tax identification, taxpayer identification, tax id, tax number, fiscal number, fiscal code, vat number, tin \\
     & Spanish & identificación fiscal, identificacion fiscal, identificación tributaria, identificacion tributaria, número fiscal, numero fiscal, nif, rfc, cif, cuit \\
     & French & numéro fiscal, numero fiscal, identification fiscale, identifiant fiscal, numéro d identification fiscale, numero d identification fiscale, numéro de tva, numero de tva \\
     & Hindi & {\devanagarifont कर पहचान}, {\devanagarifont करदाता पहचान}, {\devanagarifont टैक्स आईडी}, {\devanagarifont पैन नंबर}, {\devanagarifont स्थायी खाता संख्या} \\
     & Italian & codice fiscale, partita iva, identificazione fiscale, numero fiscale \\
     & Dutch & belastingnummer, btw nummer, btw identificatienummer, fiscaal nummer, fiscaal identificatienummer, belasting identificatienummer \\
     & Telugu & {\telugufont పన్ను గుర్తింపు}, {\telugufont పన్ను సంఖ్య}, {\telugufont ట్యాక్స్ ఐడి}, {\telugufont పాన్ నంబర్} \\
    \bottomrule
  \end{tabular}
  \caption{\textbf{Direct context words.} Words and phrases that name each category, listed after normalization. A text has direct context for a category if it contains any of that category's words in any language.}
  \label{tab:c1}
\end{table*}

\begin{table*}[tp]
  \centering\small
  \renewcommand{\thetable}{C2}\renewcommand{\theHtable}{C2}
  \begin{tabular}{@{}L{0.08\textwidth}L{0.88\textwidth}@{}}
    \toprule
    Language & Broad context words \\
    \midrule
    German & lizenz, führerschein, fuhrerschein, fahrerlaubnis, ausweis, id, identifikation, identität, identitaet, verifizieren, verifizierung, überprüfen, ueberpruefen, überprüfung, ueberpruefung, konto, dokument, nachweis, nummer, reisepass, sozial, versicherung, steuer, fiskalisch, karte, zahlung, bezahlen, kauf, transaktion, abrechnung \\
    \addlinespace[3pt]
    English & license, licence, lic, permit, id, identification, identity, identify, verify, verification, credential, account, document, record, number, passport, social, security, tax, taxpayer, fiscal, card, payment, pay, purchase, transaction, billing, checkout \\
    \addlinespace[3pt]
    Spanish & licencia, permiso, id, identificación, identificacion, identidad, verificar, verificación, verificacion, cuenta, documento, credencial, número, numero, pasaporte, social, seguridad, impuesto, fiscal, tributario, tarjeta, pago, pagar, compra, transacción, transaccion, facturación, facturacion \\
    \addlinespace[3pt]
    French & licence, permis, pièce d identité, piece d identite, identifiant, identification, identité, identite, vérifier, verifier, vérification, verification, compte, document, justificatif, numéro, numero, passeport, social, sécurité, securite, impôt, impot, fiscal, carte, paiement, payer, achat, transaction, facturation \\
    \addlinespace[3pt]
    Hindi & {\devanagarifont लाइसेंस}, {\devanagarifont लाइसेन्स}, {\devanagarifont परमिट}, {\devanagarifont आईडी}, {\devanagarifont पहचान}, {\devanagarifont सत्यापन}, {\devanagarifont सत्यापित}, {\devanagarifont खाता}, {\devanagarifont अकाउंट}, {\devanagarifont दस्तावेज}, {\devanagarifont रिकॉर्ड}, {\devanagarifont संख्या}, {\devanagarifont नंबर}, {\devanagarifont पासपोर्ट}, {\devanagarifont सामाजिक}, {\devanagarifont सुरक्षा}, {\devanagarifont कर}, {\devanagarifont टैक्स}, {\devanagarifont कार्ड}, {\devanagarifont भुगतान}, {\devanagarifont पेमेंट}, {\devanagarifont खरीद}, {\devanagarifont लेनदेन}, {\devanagarifont बिलिंग} \\
    \addlinespace[3pt]
    Italian & licenza, patente, id, identificazione, identità, identita, verificare, verifica, conto, account, documento, credenziale, numero, passaporto, sociale, sicurezza, imposta, fiscale, tributario, carta, pagamento, pagare, acquisto, transazione, fatturazione \\
    \addlinespace[3pt]
    Dutch & licentie, vergunning, rijbewijs, id, identificatie, identiteit, verifiëren, verifieren, verificatie, rekening, account, document, legitimatie, nummer, paspoort, sociaal, zekerheid, belasting, fiscaal, kaart, betaling, betalen, aankoop, transactie, facturering \\
    \addlinespace[3pt]
    Telugu & {\telugufont లైసెన్స్}, {\telugufont పర్మిట్}, {\telugufont ఐడి}, {\telugufont గుర్తింపు}, {\telugufont ధృవీకరణ}, {\telugufont ధృవీకరించండి}, {\telugufont ఖాతా}, {\telugufont అకౌంట్}, {\telugufont పత్రం}, {\telugufont డాక్యుమెంట్}, {\telugufont రికార్డ్}, {\telugufont సంఖ్య}, {\telugufont నంబర్}, {\telugufont పాస్}{\devanagarifont {\telugufont ‌}}{\telugufont పోర్ట్}, {\telugufont సామాజిక}, {\telugufont భద్రత}, {\telugufont పన్ను}, {\telugufont ట్యాక్స్}, {\telugufont కార్డ్}, {\telugufont చెల్లింపు}, {\telugufont పేమెంట్}, {\telugufont కొనుగోలు}, {\telugufont లావాదేవీ}, {\telugufont బిల్లింగ్} \\
    \bottomrule
  \end{tabular}
  \caption{\textbf{Broad context words.} Words that give looser related context, listed after normalization and shared across the five categories in Table~\ref{tab:1}.}
  \label{tab:c2}
\end{table*}

\section{Category Mappings}\label{app:D}

Tables~\ref{tab:d1} and~\ref{tab:d2} list the label mappings described in Section~\ref{sec:4.2}. NVIDIA GLiNER's label names come from its training dataset, Nemotron-PII. We prompt each GLiNER model with all of its mapped labels at once, and when several labels map to one category, we combine their scores by taking the maximum.

\begin{table*}[p]
  \centering\small
  \renewcommand{\thetable}{D1}\renewcommand{\theHtable}{D1}
  \begin{tabular*}{\textwidth}{@{\extracolsep{\fill}}lll@{}}
    \toprule
    Category & AI4Privacy label & Nemotron label(s) \\
    \midrule
    Bank Routing Code & --- & \texttt{bank\_routing\_number} \\
    Credit Card Number & \texttt{CREDITCARDNUMBER} & \texttt{credit\_debit\_card} \\
    Driver License Number & \texttt{DRIVERLICENSENUM} & --- \\
    Email Address & \texttt{EMAIL} & \texttt{email} \\
    IP Address & --- & \texttt{ipv4}, \texttt{ipv6} \\
    Passport Number & \texttt{PASSPORTNUM} & --- \\
    Phone Number & \texttt{TELEPHONENUM} & \texttt{phone\_number}, \texttt{fax\_number} \\
    Social Security Number & \texttt{SOCIALNUM} & \texttt{ssn} \\
    Tax Identification Number & \texttt{TAXNUM} & \texttt{tax\_id} \\
    \bottomrule
  \end{tabular*}
  \caption{\textbf{Benchmark label mappings.} Benchmark labels mapped to each category. A dash denotes that the category is not comparable under our methodology.}
  \label{tab:d1}
\end{table*}

\begin{table*}[p]
  \centering\small
  \renewcommand{\thetable}{D2}\renewcommand{\theHtable}{D2}
  \begin{tabular}{@{}L{0.22\textwidth}L{0.23\textwidth}L{0.24\textwidth}L{0.18\textwidth}@{}}
    \toprule
    Category & NVIDIA GLiNER & Fastino GLiNER2 & Privacy Filter \\
    \midrule
    Bank Routing Code & \texttt{bank\_routing\_number} & \texttt{routing\_number} & --- \\
    Credit Card Number & \texttt{credit\_debit\_card} & \texttt{card\_number} & --- \\
    Driver License Number & --- & \texttt{drivers\_license\_number} & --- \\
    Email Address & \texttt{email} & \texttt{email} & \texttt{private\_email} \\
    IP Address & \texttt{ipv4}, \texttt{ipv6} & \texttt{ip\_address} & --- \\
    Passport Number & --- & \texttt{passport\_number} & --- \\
    Phone Number & \texttt{phone\_number}, \texttt{fax\_number} & \texttt{phone\_number} & \texttt{private\_phone} \\
    Social Security Number & \texttt{ssn} & --- & --- \\
    Tax Identification Number & \texttt{tax\_id} & \texttt{tax\_id}, \texttt{tax\_number} & --- \\
    \bottomrule
  \end{tabular}
  \caption{\textbf{Baseline label mappings.} Baseline labels mapped to each category. A dash denotes that the category is not comparable under our methodology.}
  \label{tab:d2}
\end{table*}

\begin{table*}[p]
  \centering\small
  \renewcommand{\thetable}{E1}\renewcommand{\theHtable}{E1}
  \begin{tabular}{@{}L{0.15\textwidth}L{0.25\textwidth}L{0.14\textwidth}L{0.14\textwidth}L{0.20\textwidth}@{}}
    \toprule
    Detail & Ours & NVIDIA GLiNER & Fastino GLiNER2 & Privacy Filter \\
    \midrule
    Chunk / overlap (tokens) & 512 / 128 & 350 / 50 & 384 / 64 & 128,000 / 0 \\
    \addlinespace[3pt]
    Category score & Maximum global-classifier score & Maximum span score & Maximum span score & Maximum token score \\
    \addlinespace[3pt]
    Spans & Token-classifier spans in categories the global classifier flags & Spans above a score threshold & Spans above a score threshold & Viterbi spans in categories above a score threshold \\
    \bottomrule
  \end{tabular}
  \caption{\textbf{Model inference.} How each model chunks its input, computes category scores, and produces spans.}
  \label{tab:e1}
\end{table*}

\makeatletter
\setlength{\@dblfptop}{0pt}
\makeatother

\begin{table*}[p]
  \centering\small
  \renewcommand{\thetable}{F1}\renewcommand{\theHtable}{F1}
  \begin{tabular*}{\textwidth}{@{\extracolsep{\fill}}llcccc@{}}
    \toprule
    Dataset & Category & Ours & NVIDIA GLiNER & Fastino GLiNER2 & Privacy Filter \\
    \midrule
    AI4Privacy & Credit Card Number & \textbf{0.985 / 0.984} & 0.705 / 0.605 & 0.706 / 0.578 & --- \\
     & Driver License Number & \textbf{0.934 / 0.462} & --- & 0.809 / 0.235 & --- \\
     & Email Address & \textbf{0.999 / 1.000} & 0.992 / 0.992 & 0.993 / 0.994 & 0.992 / 0.993 \\
     & Passport Number & \textbf{0.980 / 0.822} & --- & 0.885 / 0.345 & --- \\
     & Phone Number & 0.956 / 0.877 & 0.880 / 0.835 & \textbf{0.962} / 0.926 & 0.961 / \textbf{0.968} \\
     & Social Security Number & \textbf{0.949 / 0.414} & 0.533 / 0.162 & --- & --- \\
     & Tax Identification Number & \textbf{0.899 / 0.617} & 0.664 / 0.309 & 0.580 / 0.224 & --- \\
    Nemotron & Bank Routing Code & \textbf{1.000} / 0.995 & \textbf{1.000 / 0.997} & 0.886 / 0.880 & --- \\
     & Credit Card Number & \textbf{0.999} / 0.998 & 0.998 / \textbf{0.999} & 0.989 / 0.986 & --- \\
     & Email Address & 0.999 / \textbf{0.999} & \textbf{1.000 / 0.999} & \textbf{1.000 / 0.999} & 0.989 / 0.988 \\
     & IP Address & \textbf{1.000 / 1.000} & 0.726 / 0.996 & 0.703 / 0.976 & --- \\
     & Phone Number & \textbf{0.998 / 0.998} & 0.996 / \textbf{0.998} & 0.954 / 0.955 & 0.899 / 0.902 \\
     & Social Security Number & \textbf{0.997 / 0.996} & 0.985 / 0.993 & --- & --- \\
     & Tax Identification Number & \textbf{0.989} / 0.995 & 0.941 / \textbf{0.999} & 0.962 / 0.974 & --- \\
    \bottomrule
  \end{tabular*}
  \caption{\textbf{Detection recall at 0.25\% WildChat FPR.} Each cell reports constitutional / original recall on the same benchmark examples. Thresholds are independently calibrated at the stated FPR limit. Recall is rounded to three decimal places, and the highest value in each comparison is bolded. A dash denotes that a comparison was not available under our methodology.}
  \label{tab:f1}
\end{table*}

\begin{table*}[p]
  \centering\small
  \renewcommand{\thetable}{F2}\renewcommand{\theHtable}{F2}
  \begin{tabular*}{\textwidth}{@{\extracolsep{\fill}}llcccc@{}}
    \toprule
    Dataset & Category & Ours & NVIDIA GLiNER & Fastino GLiNER2 & Privacy Filter \\
    \midrule
    AI4Privacy & Credit Card Number & \textbf{0.986 / 0.993} & 0.707 / 0.606 & 0.815 / 0.719 & --- \\
     & Driver License Number & \textbf{0.949 / 0.598} & --- & 0.870 / 0.298 & --- \\
     & Email Address & \textbf{0.999 / 1.000} & 0.994 / 0.995 & 0.997 / 0.997 & 0.994 / 0.994 \\
     & Passport Number & \textbf{0.985 / 0.881} & --- & 0.940 / 0.511 & --- \\
     & Phone Number & \textbf{0.991} / 0.960 & 0.897 / 0.858 & 0.976 / 0.951 & 0.967 / \textbf{0.976} \\
     & Social Security Number & \textbf{0.970 / 0.549} & 0.586 / 0.200 & --- & --- \\
     & Tax Identification Number & \textbf{0.953 / 0.807} & 0.674 / 0.323 & 0.678 / 0.324 & --- \\
    Nemotron & Bank Routing Code & \textbf{1.000} / 0.995 & \textbf{1.000 / 0.997} & 0.994 / 0.988 & --- \\
     & Credit Card Number & \textbf{0.999 / 0.999} & 0.998 / \textbf{0.999} & 0.993 / 0.990 & --- \\
     & Email Address & \textbf{1.000} / 0.999 & \textbf{1.000 / 1.000} & \textbf{1.000} / 0.999 & 0.990 / 0.990 \\
     & IP Address & \textbf{1.000 / 1.000} & 0.736 / 0.996 & 0.751 / 0.991 & --- \\
     & Phone Number & \textbf{0.999 / 0.999} & 0.997 / 0.998 & 0.983 / 0.983 & 0.917 / 0.920 \\
     & Social Security Number & \textbf{0.999 / 0.998} & 0.990 / 0.996 & --- & --- \\
     & Tax Identification Number & \textbf{0.996} / 0.998 & 0.950 / \textbf{0.999} & 0.983 / 0.987 & --- \\
    \bottomrule
  \end{tabular*}
  \caption{\textbf{Detection recall at 0.5\% WildChat FPR.} Each cell reports constitutional / original recall on the same benchmark examples. Thresholds are independently calibrated at the stated FPR limit. Recall is rounded to three decimal places, and the highest value in each comparison is bolded. A dash denotes that a comparison was not available under our methodology.}
  \label{tab:f2}
\end{table*}

\section{Model Inference}\label{app:E}

We evaluate the released checkpoints \nolinkurl{nvidia/gliner-PII}, \nolinkurl{fastino/gliner2-privacy-filter-PII-multi}, and \nolinkurl{openai/privacy-filter}. All detectors run in bfloat16 using the documented output categories and mappings in Appendix~\ref{app:D}. Scores repeated across chunks are merged by taking their maximum, and repeated span candidates are deduplicated after their offsets are mapped back to the complete input (Table~\ref{tab:e1}). We split long inputs into overlapping chunks that fit each model's input length. Our model uses the 512-token chunks it was trained on, NVIDIA GLiNER uses chunks below its 384-word maximum input length so that text is not truncated, and Fastino GLiNER2 uses the defaults of its long-text inference method. Privacy Filter supports a 128,000-token context, so we pass each input in whole. Its Viterbi decoder returns spans without scores, so we score each category by the maximum marginal probability of its BIOES tags across tokens and keep the decoded spans of categories above the threshold. Spans are compared after removing leading and trailing whitespace. Efficiency measurements are taken after warm-up, with the same settings for every detector.

\section{Operating-Point Sensitivity}\label{app:F}

We recalibrate every detector and category independently at each WildChat FPR limit while keeping the benchmark inputs and model scores fixed. At 0.25\% FPR, our model has the highest constitutional recall in 12 of 14 comparisons and original-label recall in 10 of 14 (Table~\ref{tab:f1}). At 0.5\%, the corresponding counts are 14 of 14 and 10 of 14 (Table~\ref{tab:f2}).

\paragraph{Label-Free Calibration.}
Because our FPR is computed against constitutional labels, we also calibrate by label-free flag rate, treating every WildChat flag as a false positive. At a 0.5\% flag rate, our model has the highest constitutional recall in 12 of 14 comparisons and original-label recall in 9 of 14, compared with 14 and 10 under labeled calibration at the same limit (Table~\ref{tab:f2}). This calibration is conservative toward higher-recall detectors, because flags on genuine identifiers consume the budget. Under our labels, phone numbers alone appear in 0.41\% of WildChat conversations. At 0.1\% flag rate, label-free calibration is not meaningful: the budget falls below the prevalence of categories such as Phone Number, and several baselines assign the maximum score (1.0 in bfloat16) to more conversations than the budget allows, so no threshold satisfies it.

\section{Training Datasets}\label{app:G}

The training corpus consists of examples from the 77 sources below. TyDi QA and Hermes Function Calling each contribute two subsets, which are listed separately.

\begin{sourceitems}
  \item AhmetHalil/prompt\_\allowbreak{}injections\\
    \href{https://huggingface.co/datasets/AhmetHalil/prompt_injections}{\mbox{\nolinkurl{https://}}\nolinkurl{huggingface.co/datasets/AhmetHalil/prompt_injections}}
  \item ai-safety-institute/AgentHarm\\
    \href{https://huggingface.co/datasets/ai-safety-institute/AgentHarm}{\mbox{\nolinkurl{https://}}\nolinkurl{huggingface.co/datasets/ai-safety-institute/AgentHarm}}
  \item alespalla/chatbot\_\allowbreak{}instruction\_\allowbreak{}prompts\\
    \href{https://huggingface.co/datasets/alespalla/chatbot_instruction_prompts}{\mbox{\nolinkurl{https://}}\nolinkurl{huggingface.co/datasets/alespalla/chatbot_instruction_prompts}}
  \item allenai/real-toxicity-prompts\\
    \href{https://huggingface.co/datasets/allenai/real-toxicity-prompts}{\mbox{\nolinkurl{https://}}\nolinkurl{huggingface.co/datasets/allenai/real-toxicity-prompts}}
  \item Anthropic/hh-rlhf\\
    \href{https://huggingface.co/datasets/Anthropic/hh-rlhf}{\mbox{\nolinkurl{https://}}\nolinkurl{huggingface.co/datasets/Anthropic/hh-rlhf}}
  \item apart/darkbench\\
    \href{https://huggingface.co/datasets/apart/darkbench}{\mbox{\nolinkurl{https://}}\nolinkurl{huggingface.co/datasets/apart/darkbench}}
  \item Ari-S-123/pii-detection-english-consolidated\\
    \href{https://huggingface.co/datasets/Ari-S-123/pii-detection-english-consolidated}{\mbox{\nolinkurl{https://}}\nolinkurl{huggingface.co/datasets/Ari-S-123/pii-detection-english-consolidated}}
  \item arielzilber/prompt-injection-benign-evaluation-framework\\
    \href{https://www.kaggle.com/datasets/arielzilber/prompt-injection-benign-evaluation-framework}{\mbox{\nolinkurl{https://}}\nolinkurl{www.kaggle.com/datasets/arielzilber/prompt-injection-benign-evaluation-framework}}
  \item arielzilber/prompt-injection-in-the-wild\\
    \href{https://www.kaggle.com/datasets/arielzilber/prompt-injection-in-the-wild}{\mbox{\nolinkurl{https://}}\nolinkurl{www.kaggle.com/datasets/arielzilber/prompt-injection-in-the-wild}}
  \item arielzilber/prompt-injection-suffix-attack\\
    \href{https://www.kaggle.com/datasets/arielzilber/prompt-injection-suffix-attack}{\mbox{\nolinkurl{https://}}\nolinkurl{www.kaggle.com/datasets/arielzilber/prompt-injection-suffix-attack}}
  \item Bogdan01m/Catch\_\allowbreak{}the\_\allowbreak{}prompt\_\allowbreak{}injection\_\allowbreak{}or\_\allowbreak{}jailbreak\_\allowbreak{}or\_\allowbreak{}benign\\
    \href{https://huggingface.co/datasets/Bogdan01m/Catch_the_prompt_injection_or_jailbreak_or_benign}{\mbox{\nolinkurl{https://}}\nolinkurl{huggingface.co/datasets/Bogdan01m/Catch_the_prompt_injection_or_jailbreak_or_benign}}
  \item Bravansky/compact-jailbreaks\\
    \href{https://huggingface.co/datasets/Bravansky/compact-jailbreaks}{\mbox{\nolinkurl{https://}}\nolinkurl{huggingface.co/datasets/Bravansky/compact-jailbreaks}}
  \item cais/wmdp\\
    \href{https://huggingface.co/datasets/cais/wmdp}{\mbox{\nolinkurl{https://}}\nolinkurl{huggingface.co/datasets/cais/wmdp}}
  \item CohereLabs/aya\_\allowbreak{}redteaming\\
    \href{https://huggingface.co/datasets/CohereLabs/aya_redteaming}{\mbox{\nolinkurl{https://}}\nolinkurl{huggingface.co/datasets/CohereLabs/aya_redteaming}}
  \item csHuang/SafeAligner\\
    \href{https://huggingface.co/datasets/csHuang/SafeAligner}{\mbox{\nolinkurl{https://}}\nolinkurl{huggingface.co/datasets/csHuang/SafeAligner}}
  \item DAMO-NLP-SG/MultiJail\\
    \href{https://huggingface.co/datasets/DAMO-NLP-SG/MultiJail}{\mbox{\nolinkurl{https://}}\nolinkurl{huggingface.co/datasets/DAMO-NLP-SG/MultiJail}}
  \item danielmiessler/SecLists (Ai/LLM\_\allowbreak{}Testing)\\
    \href{https://github.com/danielmiessler/SecLists/tree/master/Ai/LLM_Testing}{\mbox{\nolinkurl{https://}}\nolinkurl{github.com/danielmiessler/SecLists/tree/master/Ai/LLM_Testing}}
  \item darkmatter2222/NLU-Redact-PII-v1\\
    \href{https://huggingface.co/datasets/darkmatter2222/NLU-Redact-PII-v1}{\mbox{\nolinkurl{https://}}\nolinkurl{huggingface.co/datasets/darkmatter2222/NLU-Redact-PII-v1}}
  \item DataikuNLP/kiji-pii-training-data\\
    \href{https://huggingface.co/datasets/DataikuNLP/kiji-pii-training-data}{\mbox{\nolinkurl{https://}}\nolinkurl{huggingface.co/datasets/DataikuNLP/kiji-pii-training-data}}
  \item deadbits/vigil-jailbreak-ada-002\\
    \href{https://huggingface.co/datasets/deadbits/vigil-jailbreak-ada-002}{\mbox{\nolinkurl{https://}}\nolinkurl{huggingface.co/datasets/deadbits/vigil-jailbreak-ada-002}}
  \item declare-lab/CategoricalHarmfulQA\\
    \href{https://huggingface.co/datasets/declare-lab/CategoricalHarmfulQA}{\mbox{\nolinkurl{https://}}\nolinkurl{huggingface.co/datasets/declare-lab/CategoricalHarmfulQA}}
  \item declare-lab/HarmfulQA\\
    \href{https://huggingface.co/datasets/declare-lab/HarmfulQA}{\mbox{\nolinkurl{https://}}\nolinkurl{huggingface.co/datasets/declare-lab/HarmfulQA}}
  \item DhruvTre/jailbreakbench-paraphrase-2025-08\\
    \href{https://huggingface.co/datasets/DhruvTre/jailbreakbench-paraphrase-2025-08}{\mbox{\nolinkurl{https://}}\nolinkurl{huggingface.co/datasets/DhruvTre/jailbreakbench-paraphrase-2025-08}}
  \item E3-JSI/synthetic-multi-pii-ner-v1\\
    \href{https://huggingface.co/datasets/E3-JSI/synthetic-multi-pii-ner-v1}{\mbox{\nolinkurl{https://}}\nolinkurl{huggingface.co/datasets/E3-JSI/synthetic-multi-pii-ner-v1}}
  \item embedding-data/Amazon-QA\\
    \href{https://huggingface.co/datasets/embedding-data/Amazon-QA}{\mbox{\nolinkurl{https://}}\nolinkurl{huggingface.co/datasets/embedding-data/Amazon-QA}}
  \item google-research-datasets/tydiqa (primary task)\\
    \href{https://huggingface.co/datasets/google-research-datasets/tydiqa}{\mbox{\nolinkurl{https://}}\nolinkurl{huggingface.co/datasets/google-research-datasets/tydiqa}}
  \item google-research-datasets/tydiqa (secondary task)\\
    \href{https://huggingface.co/datasets/google-research-datasets/tydiqa}{\mbox{\nolinkurl{https://}}\nolinkurl{huggingface.co/datasets/google-research-datasets/tydiqa}}
  \item gorkem371/pii-intent-detection-multilingual\\
    \href{https://huggingface.co/datasets/gorkem371/pii-intent-detection-multilingual}{\mbox{\nolinkurl{https://}}\nolinkurl{huggingface.co/datasets/gorkem371/pii-intent-detection-multilingual}}
  \item GPT-NL/data-curation-modules\\
    \href{https://github.com/GPT-NL/data-curation-modules}{\mbox{\nolinkurl{https://}}\nolinkurl{github.com/GPT-NL/data-curation-modules}}
  \item grammarly/detexd-benchmark\\
    \href{https://huggingface.co/datasets/grammarly/detexd-benchmark}{\mbox{\nolinkurl{https://}}\nolinkurl{huggingface.co/datasets/grammarly/detexd-benchmark}}
  \item gretelai/gretel-pii-masking-en-v1\\
    \href{https://huggingface.co/datasets/gretelai/gretel-pii-masking-en-v1}{\mbox{\nolinkurl{https://}}\nolinkurl{huggingface.co/datasets/gretelai/gretel-pii-masking-en-v1}}
  \item gretelai/gretel-safety-alignment-en-v1\\
    \href{https://huggingface.co/datasets/gretelai/gretel-safety-alignment-en-v1}{\mbox{\nolinkurl{https://}}\nolinkurl{huggingface.co/datasets/gretelai/gretel-safety-alignment-en-v1}}
  \item gretelai/synthetic\_\allowbreak{}multilingual\_\allowbreak{}llm\_\allowbreak{}prompts\\
    \href{https://huggingface.co/datasets/gretelai/synthetic_multilingual_llm_prompts}{\mbox{\nolinkurl{https://}}\nolinkurl{huggingface.co/datasets/gretelai/synthetic_multilingual_llm_prompts}}
  \item gretelai/synthetic\_\allowbreak{}pii\_\allowbreak{}finance\_\allowbreak{}multilingual\\
    \href{https://huggingface.co/datasets/gretelai/synthetic_pii_finance_multilingual}{\mbox{\nolinkurl{https://}}\nolinkurl{huggingface.co/datasets/gretelai/synthetic_pii_finance_multilingual}}
  \item hackaprompt/hackaprompt-dataset\\
    \href{https://huggingface.co/datasets/hackaprompt/hackaprompt-dataset}{\mbox{\nolinkurl{https://}}\nolinkurl{huggingface.co/datasets/hackaprompt/hackaprompt-dataset}}
  \item HuggingFaceH4/grok-conversation-harmless\\
    \href{https://huggingface.co/datasets/HuggingFaceH4/grok-conversation-harmless}{\mbox{\nolinkurl{https://}}\nolinkurl{huggingface.co/datasets/HuggingFaceH4/grok-conversation-harmless}}
  \item JailbreakBench/JBB-Behaviors\\
    \href{https://huggingface.co/datasets/JailbreakBench/JBB-Behaviors}{\mbox{\nolinkurl{https://}}\nolinkurl{huggingface.co/datasets/JailbreakBench/JBB-Behaviors}}
  \item JailbreakV-28K/JailBreakV-28k\\
    \href{https://huggingface.co/datasets/JailbreakV-28K/JailBreakV-28k}{\mbox{\nolinkurl{https://}}\nolinkurl{huggingface.co/datasets/JailbreakV-28K/JailBreakV-28k}}
  \item jc-detoxio/ReNeLLM-Jailbreak\\
    \href{https://huggingface.co/datasets/jc-detoxio/ReNeLLM-Jailbreak}{\mbox{\nolinkurl{https://}}\nolinkurl{huggingface.co/datasets/jc-detoxio/ReNeLLM-Jailbreak}}
  \item Lakera/gandalf\_\allowbreak{}ignore\_\allowbreak{}instructions\\
    \href{https://huggingface.co/datasets/Lakera/gandalf_ignore_instructions}{\mbox{\nolinkurl{https://}}\nolinkurl{huggingface.co/datasets/Lakera/gandalf_ignore_instructions}}
  \item Lakera/gandalf\_\allowbreak{}summarization\\
    \href{https://huggingface.co/datasets/Lakera/gandalf_summarization}{\mbox{\nolinkurl{https://}}\nolinkurl{huggingface.co/datasets/Lakera/gandalf_summarization}}
  \item Lakera/mosscap\_\allowbreak{}prompt\_\allowbreak{}injection\\
    \href{https://huggingface.co/datasets/Lakera/mosscap_prompt_injection}{\mbox{\nolinkurl{https://}}\nolinkurl{huggingface.co/datasets/Lakera/mosscap_prompt_injection}}
  \item Libr-AI/do-not-answer\\
    \href{https://github.com/Libr-AI/do-not-answer}{\mbox{\nolinkurl{https://}}\nolinkurl{github.com/Libr-AI/do-not-answer}}
  \item llm-attacks/llm-attacks\\
    \href{https://github.com/llm-attacks/llm-attacks}{\mbox{\nolinkurl{https://}}\nolinkurl{github.com/llm-attacks/llm-attacks}}
  \item meta-llama/PurpleLlama\\
    \href{https://github.com/meta-llama/PurpleLlama}{\mbox{\nolinkurl{https://}}\nolinkurl{github.com/meta-llama/PurpleLlama}}
  \item microsoft/llmail-inject-challenge\\
    \href{https://huggingface.co/datasets/microsoft/llmail-inject-challenge}{\mbox{\nolinkurl{https://}}\nolinkurl{huggingface.co/datasets/microsoft/llmail-inject-challenge}}
  \item mlcommons/ailuminate\\
    \href{https://github.com/mlcommons/ailuminate}{\mbox{\nolinkurl{https://}}\nolinkurl{github.com/mlcommons/ailuminate}}
  \item Nan-Do/code-search-net-javascript\\
    \href{https://huggingface.co/datasets/Nan-Do/code-search-net-javascript}{\mbox{\nolinkurl{https://}}\nolinkurl{huggingface.co/datasets/Nan-Do/code-search-net-javascript}}
  \item Nan-Do/code-search-net-python\\
    \href{https://huggingface.co/datasets/Nan-Do/code-search-net-python}{\mbox{\nolinkurl{https://}}\nolinkurl{huggingface.co/datasets/Nan-Do/code-search-net-python}}
  \item neulab/conala\\
    \href{https://huggingface.co/datasets/neulab/conala}{\mbox{\nolinkurl{https://}}\nolinkurl{huggingface.co/datasets/neulab/conala}}
  \item NobodyExistsOnTheInternet/ToxicDPOqa\\
    \href{https://huggingface.co/datasets/NobodyExistsOnTheInternet/ToxicDPOqa}{\mbox{\nolinkurl{https://}}\nolinkurl{huggingface.co/datasets/NobodyExistsOnTheInternet/ToxicDPOqa}}
  \item NousResearch/hermes-function-calling-v1 (multi-turn)\\
    \href{https://huggingface.co/datasets/NousResearch/hermes-function-calling-v1}{\mbox{\nolinkurl{https://}}\nolinkurl{huggingface.co/datasets/NousResearch/hermes-function-calling-v1}}
  \item NousResearch/hermes-function-calling-v1 (single-turn)\\
    \href{https://huggingface.co/datasets/NousResearch/hermes-function-calling-v1}{\mbox{\nolinkurl{https://}}\nolinkurl{huggingface.co/datasets/NousResearch/hermes-function-calling-v1}}
  \item OpenAssistant/oasst\_\allowbreak{}top1\_\allowbreak{}2023-08-25\\
    \href{https://huggingface.co/datasets/OpenAssistant/oasst_top1_2023-08-25}{\mbox{\nolinkurl{https://}}\nolinkurl{huggingface.co/datasets/OpenAssistant/oasst_top1_2023-08-25}}
  \item openguardrails/OpenGuardrailsMixZh\_\allowbreak{}97k\\
    \href{https://huggingface.co/datasets/openguardrails/OpenGuardrailsMixZh_97k}{\mbox{\nolinkurl{https://}}\nolinkurl{huggingface.co/datasets/openguardrails/OpenGuardrailsMixZh_97k}}
  \item OpenSafetyLab/Salad-Data\\
    \href{https://huggingface.co/datasets/OpenSafetyLab/Salad-Data}{\mbox{\nolinkurl{https://}}\nolinkurl{huggingface.co/datasets/OpenSafetyLab/Salad-Data}}
  \item prodnull/prompt-injection-repo-dataset\\
    \href{https://huggingface.co/datasets/prodnull/prompt-injection-repo-dataset}{\mbox{\nolinkurl{https://}}\nolinkurl{huggingface.co/datasets/prodnull/prompt-injection-repo-dataset}}
  \item raayraay/privacyleak-pii\\
    \href{https://huggingface.co/datasets/raayraay/privacyleak-pii}{\mbox{\nolinkurl{https://}}\nolinkurl{huggingface.co/datasets/raayraay/privacyleak-pii}}
  \item reshabhs/SPML\_\allowbreak{}Chatbot\_\allowbreak{}Prompt\_\allowbreak{}Injection\\
    \href{https://huggingface.co/datasets/reshabhs/SPML_Chatbot_Prompt_Injection}{\mbox{\nolinkurl{https://}}\nolinkurl{huggingface.co/datasets/reshabhs/SPML_Chatbot_Prompt_Injection}}
  \item rl-actors/Jailbreak-dataset\\
    \href{https://huggingface.co/datasets/rl-actors/Jailbreak-dataset}{\mbox{\nolinkurl{https://}}\nolinkurl{huggingface.co/datasets/rl-actors/Jailbreak-dataset}}
  \item sentraio/nercat\\
    \href{https://huggingface.co/datasets/sentraio/nercat}{\mbox{\nolinkurl{https://}}\nolinkurl{huggingface.co/datasets/sentraio/nercat}}
  \item Simsonsun/JailbreakPrompts\\
    \href{https://huggingface.co/datasets/Simsonsun/JailbreakPrompts}{\mbox{\nolinkurl{https://}}\nolinkurl{huggingface.co/datasets/Simsonsun/JailbreakPrompts}}
  \item SoftMINER-Group/TechHazardQA\\
    \href{https://huggingface.co/datasets/SoftMINER-Group/TechHazardQA}{\mbox{\nolinkurl{https://}}\nolinkurl{huggingface.co/datasets/SoftMINER-Group/TechHazardQA}}
  \item surge-ai/toxicity\\
    \href{https://github.com/surge-ai/toxicity}{\mbox{\nolinkurl{https://}}\nolinkurl{github.com/surge-ai/toxicity}}
  \item Team-ACE/ToolACE\\
    \href{https://huggingface.co/datasets/Team-ACE/ToolACE}{\mbox{\nolinkurl{https://}}\nolinkurl{huggingface.co/datasets/Team-ACE/ToolACE}}
  \item teknium/GPT4-LLM-Cleaned\\
    \href{https://huggingface.co/datasets/teknium/GPT4-LLM-Cleaned}{\mbox{\nolinkurl{https://}}\nolinkurl{huggingface.co/datasets/teknium/GPT4-LLM-Cleaned}}
  \item thu-coai/Agent-SafetyBench\\
    \href{https://huggingface.co/datasets/thu-coai/Agent-SafetyBench}{\mbox{\nolinkurl{https://}}\nolinkurl{huggingface.co/datasets/thu-coai/Agent-SafetyBench}}
  \item tom-gibbs/multi-turn\_\allowbreak{}jailbreak\_\allowbreak{}attack\_\allowbreak{}datasets\\
    \href{https://huggingface.co/datasets/tom-gibbs/multi-turn_jailbreak_attack_datasets}{\mbox{\nolinkurl{https://}}\nolinkurl{huggingface.co/datasets/tom-gibbs/multi-turn_jailbreak_attack_datasets}}
  \item toxigen/toxigen-data\\
    \href{https://huggingface.co/datasets/toxigen/toxigen-data}{\mbox{\nolinkurl{https://}}\nolinkurl{huggingface.co/datasets/toxigen/toxigen-data}}
  \item TrustAIRLab/in-the-wild-jailbreak-prompts\\
    \href{https://huggingface.co/datasets/TrustAIRLab/in-the-wild-jailbreak-prompts}{\mbox{\nolinkurl{https://}}\nolinkurl{huggingface.co/datasets/TrustAIRLab/in-the-wild-jailbreak-prompts}}
  \item TrustLLM/TrustLLM-dataset\\
    \href{https://huggingface.co/datasets/TrustLLM/TrustLLM-dataset}{\mbox{\nolinkurl{https://}}\nolinkurl{huggingface.co/datasets/TrustLLM/TrustLLM-dataset}}
  \item tyqiangz/multilingual-sentiments\\
    \href{https://huggingface.co/datasets/tyqiangz/multilingual-sentiments}{\mbox{\nolinkurl{https://}}\nolinkurl{huggingface.co/datasets/tyqiangz/multilingual-sentiments}}
  \item uiuc-kang-lab/InjecAgent\\
    \href{https://github.com/uiuc-kang-lab/InjecAgent}{\mbox{\nolinkurl{https://}}\nolinkurl{github.com/uiuc-kang-lab/InjecAgent}}
  \item urchade/synthetic-pii-ner-mistral-v1\\
    \href{https://huggingface.co/datasets/urchade/synthetic-pii-ner-mistral-v1}{\mbox{\nolinkurl{https://}}\nolinkurl{huggingface.co/datasets/urchade/synthetic-pii-ner-mistral-v1}}
  \item walledai/StrongREJECT\\
    \href{https://huggingface.co/datasets/walledai/StrongREJECT}{\mbox{\nolinkurl{https://}}\nolinkurl{huggingface.co/datasets/walledai/StrongREJECT}}
  \item walledai/TDC23-RedTeaming\\
    \href{https://huggingface.co/datasets/walledai/TDC23-RedTeaming}{\mbox{\nolinkurl{https://}}\nolinkurl{huggingface.co/datasets/walledai/TDC23-RedTeaming}}
  \item xunguangwang/JailbreakGuardrailBenchmark\\
    \href{https://huggingface.co/datasets/xunguangwang/JailbreakGuardrailBenchmark}{\mbox{\nolinkurl{https://}}\nolinkurl{huggingface.co/datasets/xunguangwang/JailbreakGuardrailBenchmark}}
\end{sourceitems}

\end{document}